\documentclass[11pt]{article}
\usepackage{nodalida2017}
\usepackage{mathptmx}
\usepackage{url}
\usepackage{latexsym}
\usepackage{csquotes}
\usepackage{hyperref}
\usepackage{listings}
\usepackage{amsmath}
\usepackage{graphicx}
\usepackage{amssymb}

\newcommand{\norm}[1]{\left\lVert#1\right\rVert}

\title{Linear Ensembles of Word Embedding Models}

\author{Avo Murom\"agi \\
 University of Tartu \\
 Tartu, Estonia \\
 {\tt avom@ut.ee} \\\And
  Kairit Sirts \\
 University of Tartu \\
 Tartu, Estonia \\
 {\tt kairit.sirts@ut.ee} \\\And
 Sven Laur \\
 University of Tartu \\
 Tartu, Estonia \\
 {\tt swen@math.ut.ee} \\
}

\date{}

\begin{document}

% DOI
%\doi{10.475/123_4}

\maketitle
\begin{abstract}
This paper explores linear methods for combining several word embedding models into an ensemble. We construct the combined models using an iterative method based on either ordinary least squares regression or the solution to the orthogonal Procrustes problem.
%Word2Vec is a tool that can represent natural language words in word embeddings by producing a vector mapping to each word in the input text. Running it on the same data multiple times produces different word embeddings that are not directly comparable.

We evaluate the proposed approaches on Estonian---a morphologically complex language, for which the available corpora for training word embeddings are relatively small. We compare both combined models with each other and with the input word embedding models using synonym and analogy tests. The results show that while using the ordinary least squares regression performs poorly in our experiments, using orthogonal Procrustes to combine several word embedding models into an ensemble model leads to 7-10\% relative improvements over the mean result of the initial models in synonym tests and 19-47\% in analogy tests.

%This project presents two linear models, a solution to linear regression coefficient estimates (SLRCE) and a solution to Orthogonal Procrustes problem (SOPP), to combine several word embedding models into a single target model. Performance of target models were compared against input models in synonym and analogy tests.

%While the SLRCE perfmed poorly in all of the tests, the SOPP significantly outperformed initial models in all synonym tests and most analogy tests and as such is a promising method for combining several word embeddings.
\end{abstract}

% <concept_id>10010147.10010178.10010179.10003352</concept_id>
% <concept_desc>Computing methodologies~Information extraction</concept_desc>

% \ccsdesc[500]{Computing methodologies~Information extraction}


%
%  Use this command to print the description
%
% \printccsdesc

% We no longer use \terms command

% \keywords{TODO}

\section{Introduction}

%It is well known that using an ensemble of models often leads to better results than using a single model. This knowledge has been often exploited in various natural language processing (NLP)  systems, such as dependency  parsers \cite{} or machine translation systems \cite{} an typically an ensemble of models improves over a single model in few percentage points in accuracy.

Word embeddings---dense low-dimensional vector representations of words---have become very popular in recent years in the field of natural language processing (NLP). Various methods have been proposed to train word embeddings from unannoted text corpora \cite{mikolov:word2vec,pennington2014,alrfou2013,turian2010,levy2014}, most well-known of them being perhaps Word2Vec \cite{mikolov:word2vec}. Embedding learning systems essentially train a model from a corpus of text and the word embeddings are the model parameters. These systems contain a randomized component and so the trained models are not directly comparable, even when they have been trained on exactly the same data. This random behaviour provides an opportunity to combine several embedding models into an ensemble which, hopefully, results in a better set of word embeddings. Although model ensembles have been often used in various NLP  systems to improve the overall accuracy, the idea of combining several word embedding models into an ensemble has not been explored before.

The main contribution of this paper is to show that word embeddings can benefit from ensemble learning, too. We study two methods for combining word embedding models into an ensemble. Both methods use a simple linear transformation. First of them is based on the standard ordinary least squares solution (OLS) for linear regression, the second uses the solution to the orthogonal Procrustes problem (OPP) \cite{Schonemann:procrustes}, which essentially also solves the OLS but adds the orthogonality constraint that keeps the angles between vectors and their distances unchanged.

There are several reasons why using an ensemble of word embedding models could be useful. First is the typical ensemble learning argument---the ensemble simply is better because it enables to cancel out random noise of individual models and reinforce the useful patterns expressed by several input models. Secondly, word embedding systems require a lot of training data to learn reliable word representations. While there is a lot of textual data available for English, there are many smaller languages for which even obtaining enough plain unannotated text for training reliable embeddings is a problem. Thus, an ensemble approach that would enable to use the available data more effectively would be beneficial.

According to our knowledge, this is the first work that attempts to leverage the data by combining several word embedding models 
%that are trained on the same input data but with different initialisations 
into a new improved model. 
Linear methods for combining two embedding models for some task-specific purpose have been used previously. \newcite{mikolov2013} optimized the linear regression with stochastic gradient descent to learn linear transformations between the embeddings in two languages for machine translation.  \newcite{mogadala2016} used OPP to translate embeddings between two languages to perform cross-lingual document classification. \newcite{hamilton2016} aligned a series of embedding models with OPP to detect changes in word meanings over time. The same problem was addressed by \newcite{kulkarni2015} who aligned the embedding models using piece-wise linear regression based on a set of nearest neighboring words for each word.

Recently, \newcite{Yin2016} experimented with several methods to learn meta-embeddings by combining different word embedding sets. Our work differs from theirs in two important aspects. First, in their work each initial model is trained with a \emph{different} word embedding system and on a \emph{different} data set, while we propose to combine the models trained with the \emph{same} system and on the \emph{same} dataset, albeit using different random initialisation. Secondly, although the 1toN model proposed in \cite{Yin2016} is very similar to the linear models studied in this paper, it doesn't involve the orthogonality constraint included in the OPP method, which in our experiments, as shown later, proves to be crucial.

%Third,  combining word embeddings can be viewed as a domain adaptation approach. Consider the scenario where there are for instance pre-trained embeddings on some general domain and then a set of domain-specific embeddings are required. If the domain corpus is large enough then the simplest solution is to train new embeddings on it. However, this is not the option when the domain corpus alone is not large enough. Then one option would be to join the general and domain corpus and train new embeddings on the joint corpus. Another option would be to load the general model and continue training embeddings with the domain data. However, some embedding models such as Glove \cite{} operate with word co-occurence counts and it's not obvious how to deduce the co-occurrence counts from the general embeddings so that they could be updated with counts from domain corpus. The third option that we explore in this paper is to train a new set of embeddings from the domain corpus and then combine the general and domain vectors into an ensemble.

We conduct experiments on Estonian and construct ensembles from ten different embedding models trained with Word2Vec. We compare the initial and combined models in synonym and analogy tests and find that the ensemble embeddings combined with orthogonal Procrustes method indeed perform significantly better in both tests, leading to a relative improvement of 7-10\% over the mean result of the initial models in synonym tests and 19-47\% in analogy tests.

\section{Combining word embedding models}

%An obvious way to minimize \ref{eq:main_min} if to set $Y = 0$ and find $P_i$ so that $W_i P_i = 0$ however this is not a useful solution. TODO: bridge
A word embedding model is a matrix $W \in \mathbb{R}^{|V| \times d}$, where $|V|$ is the number of words in the model lexicon and $d$ is the dimensionality of the vectors. Each row in the matrix $W$ is the continuous representation of a word in a vector space.

%Given input matrices $W_1, \ldots, W_r$ we wanted to combine them into a target matrix $Y.$ This required translation matrices $P_1, \ldots, P_r$ to transform matrices $W_1, \ldots, W_r$, respectively, into same vectorspace. We also had a minimization goal (\ref{eq:main_min}).
Given $r$ embedding models  $W_1, \ldots, W_r$ we want to combine them into a target model $Y$. We define a linear objective function that is the sum of $r$ linear regression optimization goals:
\begin{equation}\label{eq:main_min}
J = \sum_{i=1}^r \norm{Y - W_i P_i}^2,
\end{equation}
where $P_1, \ldots, P_r$ are transformation matrices that translate  $W_1, \ldots, W_r$, respectively, into the common vector space containing $Y$.

% This means solving $r$ equations
% \begin{equation}
% Y - W_i P_i = E_i,\bigskip i = 1, \ldots, r
% \end{equation}
% so that $tr(E_i^T E) = \text{min.}$

We use an iterative algorithm to find matrices $P_1, \ldots, P_r$ and $Y$. 
%Initial $Y$ was generated by choosing each element uniformly at random from $[-1, 1).$ 
During each iteration the algorithm performs two steps:
\begin{enumerate}
\item Solve $r$ linear regression problems with respect to the current target model $Y$, which results in updated values for matrices $P_1, \dots P_r$;
\item Update $Y$ to be the mean of the translations of all $r$ models:
\end{enumerate} 
\begin{equation}
Y = \frac 1 r \sum_{i = 1}^r W_i P_i.
\end{equation}

%$P_i$ for each $W_i$ by solving $Y - W_i P_i = E_i$ for $P_i$ while minimizing $tr(E_i^T E_i)$ and then found a new 
%\begin{equation}
%Y = \frac 1 r \sum_{i = 1}^r W_i P_i.
%\end{equation}
%

This procedure is continued until the change in the average normalised residual error, computed as
%After each iteration we calculated the error using 
%
\begin{equation}\label{eq:error}
\frac 1 r \sum_{i = 0}^r \frac {\norm{Y - W_i P_i}} {\sqrt{|V| \cdot d}},
\end{equation}
will become smaller than a predefined threshold value.
%
%where $m$ and $n$ are rows and columns of the matrix $Y$. If iteration decreases the error less that a certain threshold value, then we stop iterating and return $Y, P_1, \ldots, P_r$.

%We used two different ways of calculating matrices $P_1, \ldots, P_r, Y$ during each iteration. One way was to solve ${Y - W_i P_i = 0}$ for $P_i$, $i = 1, \ldots, r$ and then update $Y$ to be the average of all $W_i P_i$. The other was using Orthogonal Procrustes \cite{Schonemann:procrustes}.
We experiment with two different methods for computing the translation matrices $P_1, \dots, P_r$. The first is based on the standard least squares solution to the linear regression problem, the second method is known as solution to the Orthogonal Procrustes problem \cite{Schonemann:procrustes}.

%\subsection{Solution to linear regression coefficient estimates (SLRCE)}
\subsection{Solution with the ordinary least squares (SOLS)}

The analytical solution for a linear regression problem $Y=P W$ for finding the transformation matrix $P$, given the input data matrix $W$ and the result $Y$ is:
\begin{equation}
P = (W^T W)^{-1} W^T Y
\end{equation}

We can use this formula to update all matrices $P_i$ at each iteration. 
%The first approach is to simply minimize $Y - W_i P_i$ for each $i = 1, \ldots, r$. This means finding left inverse of $W_i$ which is equal to $(W_i^T W_i)^{-1} W_i^T$. Knowing this we can calculate $P_i = (W_i^T W_i)^{-1} W_i^T Y$ which is more widely known as the analytical solution to linear regression using least-squares method.
% The problem with this approach is that it optimizes both $W_i P_i$ and $Y$ towards $0$ which is not a useful solution. 
The problem with this approach is that because $Y$ is also unknown and will be updated repeatedly in the second step of the iterative algorithm, the OLS might lead to solutions where both $W_i P_i$ and $Y$ are optimized towards $0$ which is not a useful solution. 
In order to counteract this effect we rescale $Y$ at the start of each iteration. This is done by scaling the elements of $Y$ so that the variance of each column of $Y$ would be equal to 1.
% i. e. we set $\Var Y_{*1} = \ldots = \Var Y_{*n} = 1.$ Pseudocode for this can be seen on listing \ref{lst:pseudoanalytical}.

%\begin{minipage}{\linewidth}
%\begin{lstlisting}[language=Python,tabsize=2,frame=single,caption={Pseudocode for minimizing $Y - WP$ using SLRCE.},captionpos=b,label={lst:pseudoanalytical}]
%# Wi is a list of m * n matrices
%minimize(W, stop_at_diff)
%	Y = m * n matrix with elements chosen 
%			uniformly at random from [-1, 1)
%	previous_error = Infinity
%	error = calculate_error(Wi, Pi, Y)
%	while |prev_error - error| > stop_at_diff:
%		previous_error = error
%		for column in Y:
%			scale column so that 
%					variance(column) = 1
%		Pi = []
%		for W in Wi:
%			P = (W.T * W)^(-1) * W.T * Y
%			Pi.append(P)
%		error = calculate_error(Wi, Pi, Y)
%	return (Y, Pi)
%\end{lstlisting}
%\end{minipage}

\subsection{Solution to the Orthogonal Procrustes problem (SOPP)}

Orthogonal Procrustes is a linear regression problem of transforming the input matrix $W$ to the output matrix $Y$ using an orthogonal transformation matrix $P$ \cite{Schonemann:procrustes}. The orthogonality constraint is specified as
\begin{equation*}
P P^T = P^T P = I \label{eq:PPTequalsPTPequalsI}\\
\end{equation*}

%The second way of calculating was using solution to Orthogonal Procrustes problem. In our case the problem can be stated as the least-squares problem of transforming a given matrix $W_i$ into a given matrix $Y$ so that the sums of squares of the residual matrix $E = W_i P_i - Y$ is a minimum \cite{Schonemann:procrustes}. Mathematically this can be stated as follows.

%\begin{align}
%W_i P_i &= Y + E,\\
%P_i P_i^T &= P_i^T P_i = I, \label{eq:PPTequalsPTPequalsI}\\
%tr(E^T E) &= \text{min}.
%\end{align}

The solution to the Orthogonal Procrustes can be computed analytically using singular value decomposition (SVD). First compute:
%For each $W_i$ this means first computing $S = W_i^T Y$, then diagonalizing 
\begin{equation*}
S = W^T Y
\end{equation*}
Then diagonalize using SVD:
\begin{align*}
S^T S &= V D_S V^T\\
S S^T &= U D_S U^T
\end{align*}
%using singular value decomposition and then
Finally compute: 
\begin{equation*}
P = U V^T
\end{equation*}
%This has to be done $r$ times for each $i = 0, \ldots, r$ within each iteration. 
This has to be done for each $P_i$ during each iteration.

This approach is very similar to SOLS. The only difference is the additional orthogonality constraint that gives a potential advantage to this method as in the translated word embeddings $W_i P_i$ the lengths of the vectors and the angles between the vectors are preserved. Additionally, we no longer need to worry about the trivial solution where $P_1, \ldots, P_r$ and $Y$ all converge towards $\mathbf{0}$.

%Notice that the approach is very similar to SLRCE. The difference here is that we have an additional restriction (\ref{eq:PPTequalsPTPequalsI}) which gives a potential advantage to this method as it implies that $P_i$-s are orthogonal matrices. Therefore linear transformations performed using $P_i$-s preserve the lengths of the vectors and angles between vectors. Also we no longer need to worry about a trivial solution where $P_1, \ldots, P_r = \mathbf{0}$ and $Y = \mathbf{0}$.

%Pseudocode can be seen on listing \ref{lst:pseudoprocrustes}. We used the implementation built in to Python's SciPy library\footnote{SciPy linalg module orthogonal\_procrustes method documentation: \url{https://docs.scipy.org/doc/scipy-0.17.0/reference/generated/scipy.linalg.orthogonal\_procrustes.html}}.

%\begin{minipage}{\linewidth}
%\begin{lstlisting}[language=Python,tabsize=2,frame=single,caption={Pseudocode for minimizing $Y - WP$ using SOPP.},captionpos=b,label={lst:pseudoprocrustes}]
%# Wi is a list of m * n matrices
%minimize(W, stop_at_diff)
%	Y = m * n matrix with elements chosen 
%			uniformly at random from [-1, 1)
%	previous_error = Infinity
%	error = calculate_error(Wi, Pi, Y)
%	while |prev_error - error| > stop_at_diff:
%		previous_error = error
%		Pi = []
%		for W in Wi:
%			P = orthogonal_procrustes(Y, W)
%			Pi.append(P)
%		error = calculate_error(Wi, Pi, Y)
%	return (Y, Pi)
%\end{lstlisting}
%\end{minipage}

%$P_i = (W_i^T W_i)^{-1} W_i^T Y$ for each $W_i$

\section{Experiments}

\begin{table}[t]
\centering
\begin{tabular}{ccccc}
\hline
& \multicolumn{2}{c}{\bf SOLS} & \multicolumn{2}{c}{\bf SOPP} \\
\bf Dim & \bf Error & \# \bf Iter & \bf Error & \# \bf Iter \\
\hline
 50 & 0.162828 & 33 & 0.200994 & 5 \\
100 & 0.168316 & 38 & 0.183933 & 5 \\
150 & 0.169554 & 41 & 0.171266 & 4 \\
200 & 0.172987 & 40 & 0.167554 & 4 \\
250 & 0.175723 & 40 & 0.164493 & 4 \\
300 & 0.177082 & 40 & 0.160988 & 4 \\
\hline
\end{tabular}
\caption{Final errors and the number of iterations until convergence for both SOLS and SOPP. The first column shows the embedding size.}\label{error_converge_table}. 
\end{table}

We tested both methods on a number of Word2Vec models \cite{mikolov:word2vec} trained on the Estonian Reference Corpus.\footnote{\url{http://www.cl.ut.ee/korpused/segakorpus}} Estonian Reference Corpus is the largest text corpus available for Estonian. Its size is approximately 240M word tokens, which may seem like a lot but compared to for instance English Gigaword corpus, which is often used to train word embeddings for English words and which contains more than 4B words, it is quite small. 
All models were trained using a window size 10 and the skip-gram architecture. We experimented with models of 6 different embedding sizes: 50, 100, 150, 200, 250 and 300. For each dimensionality we had 10 models available. The number of distinct words in each model is 816757.

During training the iterative algorithm was run until the convergence threshold $th=0.001$ was reached. The number of iterations needed for convergence for both methods and for models with different embedding size are given in Table~\ref{error_converge_table}. It can be seen that the convergence with SOPP took significantly fewer iterations than with SOLS. This difference is probably due to two aspects: 1) SOPP has the additional orthogonality constraint which reduces the space of feasible solutions; 2) although SOLS uses the exact analytical solutions for the least squares problem, the final solution for Y does not move directly to the direction pointed to by the analytical solutions due to the variance rescaling.
% For each dimension both iterative algorithms were run until error converged (the error decreased by less than 0.001 after iteration). The error was calculated according to equation (\ref{eq:error}). All matrices $P_i$ were $n \times n$ matrices, where $n$ is the dimensionality of a word vector. $Y$ and $W$ had the same dimensions. The number of iterations needed for the error to converge and the final errors can be seen in table~\ref{error_converge_table}.

\section{Results} \label{section:results}

We evaluate the goodness of the combined models using synonym and analogy tests. 
%Analogy tests are often used to evaluate word embeddings \cite{schnabel2015}. 
%Finally, we evaluate the combined models in synonym and analogy tests.
%Then it was compared how different methods affect similarities of words. Two different tests were made to evaluate the goodness of described methods. First test was to compare how synonym similarities rank in target models and the second test was to compare how well target models can answer analogy questions.

%Both methods were tested on a number of Word2Vec models all trained on the Estonian Reference Corpus. All models were pre-trained using window size 10 and a skip-gram architecture. Models with 6 different dimensions were used: 50, 100, 150, 200, 250 and 300. For each dimensionality we had 10 models available.

\subsection{Synonym ranks}

\begin{table}[t]
\centering
\begin{tabular}{rccccc}
\hline
\bf Dim & \bf SOLS & \bf SOPP & \bf Mean $W$ \\
\hline
50 & 70098 & \bf 38998 & 41933 \\
100 & 68175 & \bf 32485 & 35986 \\
150 & 73182 & \bf 30249 & 33564 \\
200 & 73946 & \bf 29310 & 32865 \\
250 & 75884 & \bf 28469 & 32194 \\
300 & 77098 & \bf 28906 & 32729 \\
\hline
Avg & 73064 & \bf 31403 & 34879 \\
\hline
\end{tabular}
\caption{Average mean ranks of the synonym test, smaller values are better. The best result in each row is in bold. All differences are statistically significant: with $p < 2.2 \cdot 10^{-16}$ for all cases.}\label{tbl:syntest}
\end{table}

\begin{figure*}[ht]
\centering
\includegraphics[width=0.9\linewidth]{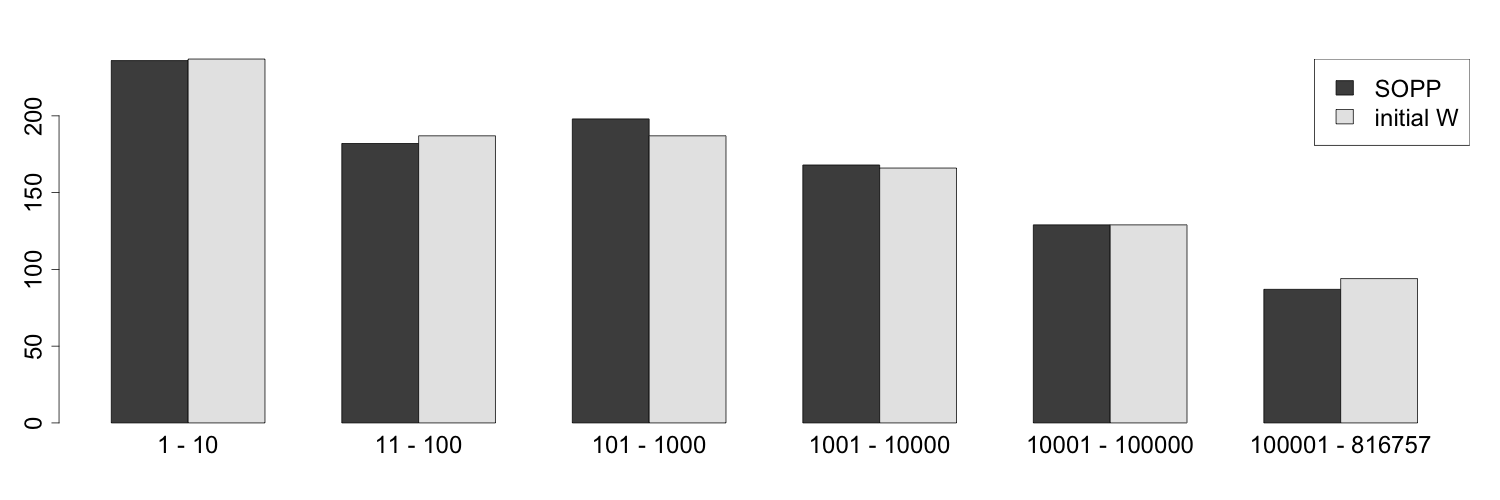}
\caption{Histogram of the synonym ranks of the 100 dimensional vectors. Dark left columns show the rank frequencies of the SOPP model, light right columns present the rank frequencies of one of the initial models.}
\label{fig:synonym_ranks}
\end{figure*}

One of the common ways to evaluate word embeddings is to use relatedness datasets to measure the correlation between the human and model judgements \cite{schnabel2015}. In those datasets, there are word pairs and each pair is human annotated with a relatedness score. The evaluation is then performed by correlating the cosine similarities between word pairs with the relatedness scores. As there are no annotated relatedness datasets for Estonian, we opted to use a synonym test instead. We rely on the assumption that the relatedness between a pair of synonyms is high and thus we expect the cosine similarity between the synonymous words to be high as well.

%Authors also measured how proposed methods maintain the rank of synonyms. For this a synonym for a number of words was obtained. Synonyms are downloaded from EKI synonym dictionary\footnote{The Institute of the Estonian Language, \url{http://www.eki.ee/dict/sys/}}. 
We obtained the synonyms from the Estonian synonym dictionary.\footnote{The Institute of the Estonian Language, \url{http://www.eki.ee/dict/sys/}}
%Each word in the vocabulary was queried and if an entry which was an exact match was found, then the first synonym offered by the dictionary was kept. If the offered synonym didn't exist in our vocabulary, then the entry was ignored. 
We queried each word in our vocabulary and when the exact match for this word was found then we looked at the first synonym offered by the dictionary. If this synonym was present in our vocabulary then the synonym pair was stored.
%In this manner a total of 7579 words were matched to a synonym. 
In this manner we obtained a total of 7579 synonym pairs.
We ordered those pairs according to the frequency of the first word in the pair and chose the 1000 most frequent words with their synonyms for the synonym test.
%1000 words with the highest frequency in the initial models were eventually used with their synonym in synonym ranking test.

%Then for each of those 1000 words cosine similarities to every other word in the vocabulary was computed, those similarities were ordered in descending order and the rank of the synonym in the resulting list was found. 
For each first word in the synonym pair, we computed its cosine similarity with every other word in the vocabulary, ordered those similarities in the descending order and found the rank of the second word of the synonym pair in this resulting list. Then we computed the mean rank over all 1000 synonym pairs.
%These steps were performed on both target models $Y_{SLRCE}$ and $Y_{SOPP}$ and also on all input models $W_i$ resulting in 1000 ranks for each model. Finally mean of the rank in each target matrix was found and also mean of the mean ranks of input models was computed.
We performed these steps on both types of combined models--- $Y_{SOLS}$ and $Y_{SOPP}$--- and also on all input models $W_i$.
%Finally mean of the rank in each target matrix was found and also mean of the mean ranks of input models was computed.
Finally we also computed the mean of the mean ranks of all 10 input models.

The results as shown in Table \ref{tbl:syntest} reveal that the synonym similarities tend to be ranked lower in the combined model obtained with  SOLS when compared to the input models. SOPP, on the other hand, produces a combined model where the synonym similarities are ranked higher than in initial models. This means that the SOPP combined models pull the synonymous words closer together than they were in the initial models. The differences in mean ranks were tested using paired Wilcoxon signed-rank test at 95\% confidence level and the differences were statistically significant with $p$-value being less than $2.2 \cdot 10^{-16}$ in all cases. In overall, the SOPP ranks are on average 10\% lower than the mean ranks of the initial models. The absolute improvement on average between SOPP and mean of $W$ is 3476.

\begin{table*}[t]
\setlength\tabcolsep{5.1pt}
\centering
\begin{tabular}{rccccc|ccccc}
\hline
& \multicolumn{5}{c|}{\bf Hit@1} & \multicolumn{5}{c}{\bf Hit@10} \\
\hline
\bf Dim & \bf SOLS & \bf SOPP & \bf Mean $W$ & \bf Min $W$ & \bf Max $W$ & \bf SOLS & \bf SOPP & \bf Mean $W$ & \bf Min $W$ & \bf Max $W$\\
\hline
 50 & 0.058 & \bf 0.193 & 0.144 & 0.124 & 0.170 & 0.158 & \bf 0.390 & 0.329 & 0.305 & 0.347 \\
100 & 0.116 & \bf 0.255 & 0.185 & 0.170 & 0.197 & 0.239 & \bf 0.475 & 0.388 & 0.371 & 0.409 \\
150 & 0.085 & \bf 0.278 & 0.198 & 0.170 & 0.228 & 0.224 & \bf 0.502 & 0.398 & 0.378 & 0.417 \\
200 & 0.066 & \bf 0.290 & 0.197 & 0.178 & 0.224 & 0.205 & \bf 0.541 & 0.408 & 0.390 & 0.425 \\
250 & 0.093 & \bf 0.282 & 0.200 & 0.181 & 0.224 & 0.193 & \bf 0.517 & 0.406 & 0.394 & 0.421 \\
300 & 0.069 & \bf 0.286 & 0.197 & 0.162 & 0.228 & 0.212 & \bf 0.533 & 0.401 & 0.359 & 0.440 \\
\hline
Avg & 0.081 & \bf 0.264 & 0.187 & & & 0.205 & \bf 0.493 & 0.388\\
\end{tabular}
\caption{Hit@1 and Hit@10 accuracies of the analogy test. SOLS and SOPP columns show the accuracies of the combined models. Mean $W$, Min $W$ and Max $W$ show the mean, minimum and maximum accuracies of the initial models $W_i$, respectively. The best accuracy among the combined models and the mean of the initial models is given in bold. The last row shows the average accuracies over all embedding sizes.}\label{tbl:analogy_test}
\end{table*}

Although we assumed that the automatically extracted synonym pairs should be ranked closely together, looking at the average mean ranks in Table~\ref{tbl:syntest} reveals that it is not necessarily the case---the average rank of the best-performing SOPP model is over 31K. In order to understand those results better we looked at the rank histogram of the SOPP model and one of the initial models, shown on Figure~\ref{fig:synonym_ranks}. Although the first bin covering the rank range from 1 to 10 contains the most words for both models and the number of synonym pairs falling to further rank bins decreases the curve is not steep and close to 100 words (87 in case of SOPP and 94 in case of the initial model) belong to the last bin counting ranks higher than 100000. Looking at the farthest synonym pairs revealed that one word in these pairs is typically polysemous and its sense in the synonym pair is a relatively rarely used sense of this word, while there are other more common senses of this word with a completely different meaning. We give some examples of such synonym pairs:
\begin{itemize}
\item \textbf{kaks} (\emph{two}) - \textbf{puudulik} (\emph{insufficient}): the sense of this pair is the insufficient grade in high school, while the most common sense of the word \textbf{kaks} is the number \emph{two};
\item \textbf{ida} (\emph{east}) - \textbf{ost} (loan word from German also meaning \emph{east}): the most common sense of the word \textbf{ost} is \emph{purchase};
\item \textbf{rubla} (\emph{rouble}) - \textbf{kull} (\emph{bank note in slang}): the most common sense of the word \textbf{kull} is \emph{hawk}. 
\end{itemize}

\subsection{Analogy tests}

Analogy tests are another common intrinsic method for evaluating word embeddings \cite{mikolov2014-linguistic-regularities}.
% Last test to compare models $Y_{SOPP}$ and $Y_{SLRCE}$ with input models was testing how accurately they can answer analogy questions. 
A famous and typical example of an analogy question is \enquote{a man is to a king like a woman is to a \_\_\_?}. 
The correct answer to this question is \enquote{queen}. 

For an analogy tuple $a: b, x:y$ ($a$ is to $b$ as $x$ is to $y$) the following is expected in an embedding space to hold:
%In a good word embedding it is expected for the followoing to hold:
\begin{equation*}
\mathbf{w}_b - \mathbf{w}_a + \mathbf{w}_x \approx \mathbf{w}_y,
\end{equation*}
where the vectors $\mathbf{w}$ are word embeddings. For the above example with \enquote{man}, \enquote{king}, \enquote{woman} and \enquote{queen} this would be computed as:
\begin{equation*}
\mathbf{w}_{king} - \mathbf{w}_{man} + \mathbf{w}_{woman} \approx \mathbf{w}_{queen}
\end{equation*}
%\[vec("king") - vec("man") + vec("woman") \approx vec("queen").\]

Given the vector representations for the three words in the analogy question---$\mathbf{w}_a$, $\mathbf{w}_b$ and $\mathbf{w}_x$---the goal is to maximize \cite{mikolov:word2vec}
\begin{equation}\label{eq:analogy_cos}
cos(\mathbf{w}_y, \mathbf{w}_b - \mathbf{w}_a + \mathbf{w}_x)
\end{equation}
over all words $y$ in the vocabulary.

We used an Estonian analogy data set with 259 word quartets. Each quartet contains two pairs of words. 
The word pairs in the data set belong into three different groups where the two pairs contain either:
%For testing, authors had 259 quartets of words for analogy testing. Each quartet contained two pairs of words. All words were from Estonian language and divided into 3 different groups where each pair contained

\begin{itemize}
	\item a positive and a comparative adjective form, e.g. \textit{pime} : \textit{pimedam}, \textit{j\~oukas} : \textit{j\~oukam} (in English \mbox{ \textit{dark} : \textit{darker},  \textit{wealthy} : \textit{wealthier}});
	\item the nominative singular and plural forms of a noun, e.g. \textit{vajadus} : \textit{vajadused}, \textit{v\~oistlus} : \textit{v\~oistlused} (in English \textit{need} : \textit{needs} , \textit{competition} : \textit{competitions});
	\item The lemma and the 3rd person past form of a verb, e.g. \textit{aitama} : \textit{aitas},  \textit{katsuma} : \textit{katsus} 
	(in English \textit{help} : \textit{helped}, \textit{touch} : \textit{touched}).
\end{itemize}

We evaluate the results of the analogy test using prediction accuracy. A prediction is considered correct if and only if the vector $\mathbf{w}_y$ that maximizes (\ref{eq:analogy_cos}) represents the word expected by the test case. We call this accuracy Hit@1. Hit@1 can be quite a noisy measurement as there could be several word vectors in a very close range to each other competing for the highest rank. Therefore, we also compute Hit@10, which considers the prediction correct if the word expected by the test case is among the ten closest words. 
As a common practice, the question words represented by the vectors $\mathbf{w}_a$, $\mathbf{w}_b$ and $\mathbf{w}_x$ were excluded from the set of possible predictions.
%We computed the accuracies for both combined models and for all initial models $W_i$.
%Accuracies were computed by performing each test case on both target models $Y_{SLRCE}$ and $Y_{SOPP}$ and all initial models $W_i.$ 
%The accuracy was calculated by dividing the number correctly made predictions with a total number of predictions. 

The Hit@1 and Hit@10 results in Table~\ref{tbl:analogy_test} show similar dynamics: combining models with SOPP is much better than SOLS in all cases. The SOPP combined model is better than the mean of the initial models in all six cases. Furthermore, it is consistently above the maximum of the best initial models.
The average accuracy of SOPP is better than the average of the mean accuracies of initial models by 41\%, relatively (7.7\% in absolute) in terms of Hit@1 and 27\% relatively (10.5\% in absolute) in terms of Hit@10.

%Another version of the analogy test was performed which considered the prediction to be a hit if the correct word appeared in top 10 predictions i. e. its vector belonged to 10 most similar vectors to a vector $v - u + u^*.$
%Similar dynamics can be observed about the Hit@10 results in Table~\ref{tbl:analogy_test_10}: the SOLS combined model performs is worse than SOPP and all input models in all cases. 
%$Y_{SOPP}$ was much better than in previous test: it had higher accuracy than the average accuracy of input models and in 4 cases out of 6 it was above the 95\% confidence bound.
%The accuracy of the SOPP combined model is better than the best accuracy of the initial model in all cases. Furthermore, it is consistently above the 95\% confidence bound. On average, the relative improvement of the SOPP model over the mean of the initial models is 27\% (10.5\% in absolute).

\section{Analysis}

In order to gain more understanding how the words are located in the combined model space in comparison to the initial models we performed two additional analyses.
First, we computed the distribution of mean squared errors of the words to see how the translated embeddings scatter around the word embedding of the combined model. 
Secondly,  we looked at how both of the methods affect the pairwise similarities of words.

\subsection{Distribution of mean squared distances}

%Running either SLRCE or SOPP yields a target matrix $Y$ and translation matrices $P_i, i=1, 2, \ldots, r$ where $r$ is the number of input matrices. Given three matrices $Y$, $W_i$ and $P_i$ a vector of squared errors $e_i$ can be computed by finding the squared Euclidean norm of each row in matrix $Y - W_i P_i.$ Then a vector of mean squared errors can be found by calculating $e = \frac 1 r \sum_{i = 0}^r e_i.$

\begin{figure}[t]
\centering
\includegraphics[width=0.9\linewidth]{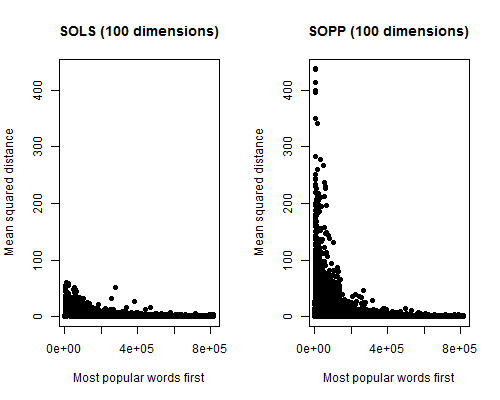}
\caption{Mean squared distances describing the scattering of the translated word embeddings around the combined model embedding for every word in the vocabulary. The words in the horizontal axis are ordered by the frequency with most frequent words plotted first.}
\label{fig:word_distances}
\end{figure}

We computed the squared Euclidean distance for each word in vocabulary between the combined model $Y$ and all the input embedding models. The distance $e_{ij}$ for a $j$th word and the $i$th input model is:
\begin{equation*}
d_{ij} = \|Y_j - T_{ij}\|^2,
\end{equation*}
where $T_i = W_i P_i$ is the $i$th translated embedding model. Then we found the mean squared distance for the $j$th word by calculating:
\begin{equation*}
d_j = \frac 1 r \sum_{i = 0}^r d_{ij}
\end{equation*}
 %$e = \frac 1 r \sum_{i = 0}^r e_i.$

%The resulting distribution can be seen on figure~\ref{fig:word_distances}.
These distances are plotted on Figure~\ref{fig:word_distances}. The words on the horizontal axis are ordered by their frequency---the most frequent words coming first. We show these results for models with 100 dimensions but the results with other embedding sizes were similar. 

Notice that the distances for less frequent words are similarly small for both SOLS and SOPP methods.
%Errors seem to be low for infrequent words and this is similar for both methods SLRCE and SOPP.
%However distribution of errors for frequent words is quite different: while the errors go up in both methods they go much higher for SOPP approach. 
However, the distribution of distances for frequent words is quite different---while the distances go up with both methods, the frequent words are much more scattered when using the SOPP approach.

\begin{figure}[t]
\centering
\includegraphics[width=0.9\linewidth]{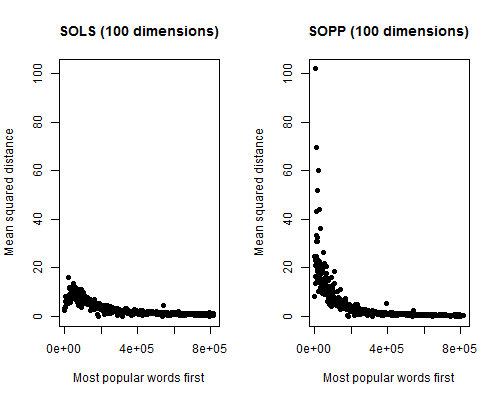}
\caption{Mean squared distances describing the scattering of the translated word embeddings around the combined model embedding for a random sample of 1000 words. The words in the horizontal axis are ordered by the frequency with most frequent words plotted first.}
\label{fig:word_distances_sample}
\end{figure}

Figure~\ref{fig:word_distances_sample} shows the mean squared distances of a random sample of 1000 words. 
These plots reveal another difference between the SOLS and SOPP methods. 
While for SOPP, the distances tend to decrease monotonically with the increase in word frequency rank, with SOLS the distances first increase and only then they start to get smaller.
%While for both methods the distances of more frequent words tend to be larger, with SOLS the distances initially have a growing trend as the word frequency goes down---with SOPP this trend 
%These plots reveal a difference in error distributions: while for both methods errors of more frequent words tend to be higher, this rule does not apply SLRCE when looking at most frequent words. In the plot for SLRCE one can see that initially errors have a growing trend as the word frequency goes down. In case of SOPP this behavior is trend does not seem to exist.

\begin{figure*}[t]
\centering
\includegraphics[width=0.9\linewidth]{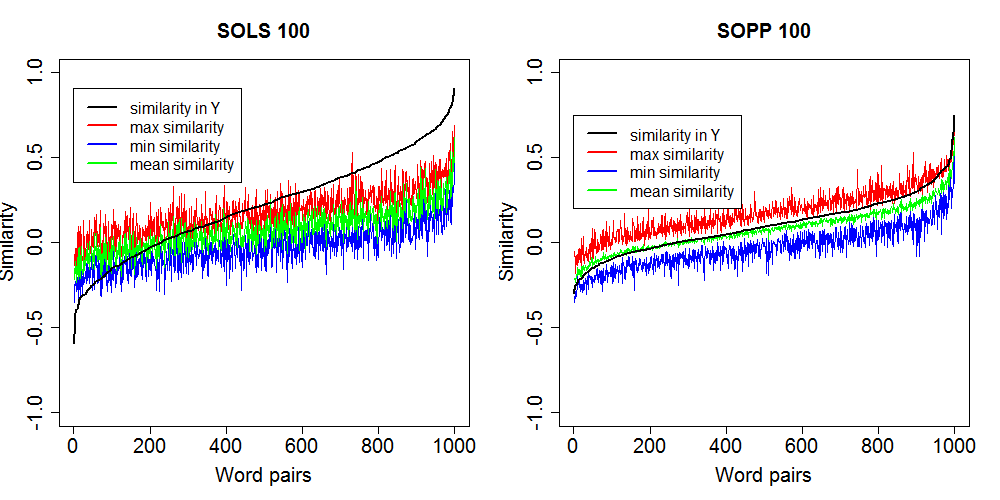}
\caption{Cosine similarities of 1000 randomly chosen word pairs ordered by their similarity in the combined model $Y$. Red, blue and green bands represent the maximum, mean and minimum similarities in the initial  models, respectively.}
\label{fig:random_similarities}
\end{figure*}

Our vocabulary also includes punctuation marks and function words, which are among the most frequent tokens and which occur in many different contexts.
%When inspecting the most frequent words they also include punctuation marks and function words which can occur in many contexts. 
Thus, the individual models have a lot of freedom to position them in the word embedding space. The SOLS combined model is able to bring those words more close to each other in the aligned space, while SOPP has less freedom to do that because of the orthogonality constraint. When looking at the words with largest distances under SOPP in the 1000 word random sample then we see that the word with the highest mean squared distance refers to the proper name of a well-known Estonian politician who has been probably mentioned often and in various contexts in the training corpus. Other words with a large distance in this sample include for instance a name of a month and a few quantifying modifiers.

\subsection{Word pair similarities}

In this analysis we looked at how the cosine similarities between pairs of words change in the combined model compared to their similarities in the input embedding models.
% In order to compare how cosine similarities between words are affected in $Y_{SLRCE}$ and $Y_{SOPP}$ a total of 1000 pairs of words were randomly chosen from the vocabulary. 
For that, we chose a total of 1000 word pairs randomly from the vocabulary.
For each pair we calculated the following values:
\begin{itemize}
	\item cosine similarity under the combined model;
	\item maximum and minimum cosine similarity in the initial models~$W_i$;
	\item mean cosine similarity over the initial models~$W_i$.
\end{itemize}

The results are plotted in Figure~\ref{fig:random_similarities}. 
These results are obtained using the word embeddings with size 100, using different embedding sizes revealed the same patterns.
%The results with different embedding sizes were similar.
%The resulting similarities were put on two plots: one plot for each method. 
%Word pairs were order in the ascending order of their similarities in $Y.$ 
In figures, the word pairs are ordered on the horizontal axis in the ascending order of their similarities in the combined model~$Y$.
%The result can be seen on figure~\ref{fig:random_similarities} and they reveal that 1) words that are similar in initial models $W_i$ are even more similar in target models $Y$ and 2) distant words in initial models are even more distant in target models. 

The plots reveal that 1) the words that are similar in initial models $W_i$ are even more similar in the combined model $Y$; and 2) distant words in initial models become even more distant in the combined model. 
Although these trends are visible in cases of both SOLS and SOPP, this behaviour of the combined models to bring more similar words closer together and place less similar words farther away is more emphasized in the combined model obtained with SOLS.
%The previous is true for both SLRCE and SOPP, however the difference to initial models is much more emphasized in a target model produced via SLRCE.

In Figure~\ref{fig:random_similarities}, the red, green and blue \enquote{bands}, representing the maximum, mean and minimum similarities of the initial models, respectively, are wider on the SOLS plot. This indicates that SOPP preserves more the original order of word pairs in terms of their similarities. However, some of this difference may be explained by the fact that SOPP has an overall smaller effect on the similarity compared to SOLS, which is due to the property of SOPP to preserve the angles and distances between the vectors during the transformation.

\section{Discussion and future work}

From the two linear methods used to combine the models, SOPP was performing consistently better in both synonym and analogy tests. Although, as shown in Figures~\ref{fig:word_distances} and \ref{fig:word_distances_sample}, the word embeddings of the aligned initial models were more closely clustered around the embeddings of the SOLS combined model, this seemingly better fit is obtained at the cost of distorting the relations between the individual word embeddings. Thus, we have provided evidence that adding the orthogonality constraint to the linear transformation objective is important to retain the quality of the translated word embeddings. This observation is relevant both in the context of producing model ensembles as  well as in other contexts where translating one embedding space to another could be relevant, such as when working with semantic time series or multilingual embeddings.

In addition to combining several models trained on the same dataset with the same configuration as demonstrated in this paper, there are other possible use cases for the model ensembles which could be explored in future work. For instance, currently all our input models had the same dimensionality and the same embedding size was also used in the combined model. In future it would be interesting to experiment with combining models with different dimensionality, in this way marginalising out the embedding size hyperparameter. 

Our experiments showed that the SOPP approach performs well in both synonym and analogy tests when combining the models trained on the relatively small Estonian corpus. In future we plan to conduct similar experiments on more languages that, similar to Estonian, have limited resources for training reliable word embeddings.

Another idea would be to combine embeddings trained with different models. As all word embedding systems learn slightly different embeddings, combining for instance Word2Vec \cite{mikolov:word2vec}, Glove \cite{pennington2014} and dependency based vectors \cite{levy2014} could lead to a model that combines the strengths of all the input models. \newcite{Yin2016} demonstrated that the combination of different word embeddings can be useful. However, their results showed that the model combination is less beneficial when one of the input models (Glove vectors in their example) is trained on a huge text corpus. Thus, we predict that the ensemble of word embeddings constructed based on different embedding models also has the most effect in the setting of limited training resources.

Finally, it would be interesting to explore the domain adaptation approach by combining for instance the embeddings learned from the large general domain with the embeddings trained on a smaller domain specific corpus. This could be of interest because there are many pretrained word embedding sets available for English that can be freely downloaded from the internet, while the corpora they were trained on (English Gigaword, for instance) are not freely available. The model combination approach would enable to adapt those embeddings to the domain data by making use of the pretrained models.

\section{Conclusions}

%The goal of the project was to implement two different methods SLRCE and SOPP to produce a target model $Y$ by combining input models $W_1, \ldots, W_r.$ Target models were compared to input models. One noticeable difference was that words similar in input models were even more similar in target models and words distant in input models were even more distant in target models. Also, given two pairs of words the probability that the pair with higher similarity still has higher similarity in target model is higher for SOPP than for SLRCE. This indicates that SOPP preserves the order of similarities better than SLRCE.
Although model ensembles have been often used to improve the results of various natural language processing tasks, the ensembles of word embedding models have been rarely studied so far. 
Our main contribution in this paper was to combine several word embedding models trained on the same dataset via linear transformation into an ensemble and demonstrate the usefulness of this approach experimentally.

We experimented with two linear methods to combine the input embedding models---the ordinary least squares solution to the linear regression problem and the orthogonal Procrustes which adds an additional orthogonality constraint to the least squares objective function. Experiments on synonym and analogy tests on Estonian showed that the combination with orthogonal Procrustes was consistently better than the ordinary least squares, meaning that preserving the distances and angles between vectors with the orthogonality constraint is crucial for model combination. Also, the orthogonal Procrustes combined model performed better than the average of the individual initial models in all synonym tests and analogy tests suggesting that combining several embedding models is a simple and useful approach for improving the quality of the word embeddings.

%Synonym rank tests revealed than synonyms ranked considerably higher in target models produced by SOPP when compared to input models. SLRCE, however, produced target models that ranked synonyms a lot lower that they were ranked in input models. These differences were statistically significant leading to a conclusion that a target model SOPP was an improvement over input models and the opposite was true for SLRCE.

%Results of analogy tests were similar. SLRCE produced target models consistently much worse than input models. In recall at 1 analogy tests SOPP performed inconsistently by producing an improvement in 3 cases out of 6 and deterioration in 3 other cases. Recall at 10 analogy tests were much better for SOPP, by producing 4 target models out of 6 that were significantly better than initial models with accuracies above 95\% confidence interval. In two other test cases accuracies were near bound of 95\% confidence interval.

%The results of the project show that SOPP produces better models than SLRCE. Target models from the latter are worse than input models. Synonym rank and analogy tests produced evidence that SOPP produces an improvement over input models making it a viable option to combine multiple Word2Vec models into a single model.

%ACKNOWLEDGMENTS are optional
\section*{Acknowledgments}

%Special thanks to Sven Laur for providing the initial problem statement, input models and analogy tests and for his advice on how to approach the problem.
We thank Alexander Tkachenko for providing the pretrained input models and the analogy test questions. We also thank the anonymous reviewers for their helpful suggestions and comments.

\bibliographystyle{acl}
\bibliography{biblio}  % sigproc.bib is the name of the Bibliography in this case
% ACM needs 'a single self-contained file'!
\end{document}